# Leveraging Energy Features for Surface Classification with Deep Learning: A Comparative Analysis Across Three Independent Datasets

Alexander Belyaev, Oleg Kushnarev

***Abstract*—The energy-based method remains a comparatively underexamined approach for surface classification in mobile robotics, despite promising results in constrained environments. This study evaluated the viability of using energy-derived features as either a standalone classification modality or as supplementary input to inertial data. A comprehensive evaluation was conducted across three publicly available datasets, comparing the performance of modern deep learning architectures including recurrent neural networks, convolutional neural networks, encoder-only transformers, and Mamba state-space models, under automated hyperparameter tuning and input sequence length optimization. The models achieved higher accuracy than previously reported values on all evaluated datasets, with the convolutional neural network yielding the highest overall performance. When relying exclusively on energy-based features, the models attained classification accuracies in the range of 85-90%, approximately 5-10% lower than those achieved when combined with inertial features (96-99%). Augmenting inertial data with energy features resulted in a consistent mean accuracy improvement of 1-2%. These findings indicate that classifiers relying solely on energy features offer sufficient accuracy for standalone deployment, while also providing a consistent gain when used in combination with other sensing modalities.**

## I. INTRODUCTION

The application of wheeled mobile robots is rapidly expanding beyond indoor environments to encompass complex, heterogeneous outdoor spaces, including fields such as logistics, agriculture, planetary exploration, and disaster response. This transition requires reliable autonomous operation on diverse and often unpredictable terrain. A critical but not yet fully resolved challenge lies in the diverse influence of the underlying surface on the robotic platform.

This influence manifests itself primarily in two ways. The first one is through direct effects on external sensors, such as dust, vibrations, and lighting conditions affecting cameras and lidars. The second one, which is more important, is through interaction with the robot's actuators.

Authors contributed equally. *Corresponding author: Alexander Belyeaev*

Alexander Belyaev and Oleg Kushnarev are with National Research Tomsk Polytechnic University, Tomsk, Russia (e-mail: asb22@tpu.ru, okushnarev@{tpu.ru, icloud.com}).

For general-purpose robots, such as autonomous delivery platforms, this leads to suboptimal performance, increased energy consumption and wear, and deviation from the intended path. The consequences are critical to mission success for robots operating in extreme conditions, whether off-road vehicles or search-and-rescue operations. Unaccounted changes in surface properties directly affect a robot's mobility, and misjudging them can lead to immobilization, mechanical failure, and, ultimately, mission failure.

Consequently, accurate real-time estimation of terrain parameters, such as traction, angle of internal friction, roughness, is important for high-level navigation functions, such as path planning, motion control, and traversability assessment. However, the estimation of these terrain parameters alone is insufficient for predicting the resultant vehicle-terrain interaction. The observed negative effects stem from both terrain and robot properties, including wheel geometry, suspension, vehicle weight, speed, and the trajectory itself. This complex interplay makes constructing general analytical or empirical models extremely difficult.

As a pragmatic and effective compromise, the robotics community often reduces this problem to classifying surfaces into a discrete set of semantically meaningful categories, such as asphalt, grass, gravel, sand, etc. Each class is associated with pre-learned or model-based parameters for navigation.

Surface classification methods are typically divided into two categories: exteroceptive and proprioceptive. The basis for this distinction lies in the nature of the information utilized [1]. Exteroceptive methods operate on data obtained remotely, while proprioceptive methods utilize information obtained through contact with the surface. The first group uses cameras, lidars, radars and provides, for the most part, only the visual component of surface properties. While the other group allows determining both surface properties, such as texture and friction coefficient, and characteristics of direct influence on the robot, such as traction and slip coefficients.

The utilization of exteroceptive sensors in conjunction with deep learning methodologies has recently emerged as a highly effective approach for achieving high-accuracy surface segmentation and detection [2-8]. However, to determine surface properties, such methods are combined with proprioceptive sensors. For example, [2] proposed the integration of a stereo camera and an inertial measurement unit (IMU). The surface is classified based on the inertial module's readings when the robot encounters it. Concurrently, data from the stereo camera is used to form and store a visual representation of the surface. An extension of this approach, incorporating the ability to learn previously unknown classes, is presented in [3]. In [4], the authors used visual and acoustic information processed by an encoder-decoder convolutional neural network and achieved 95% classification accuracy and 57% mIoU segmentation metric. A similar approach was used in [5], but due to a larger set of surfaces, the average accuracy drops to 85%. In [6], the authors, using visual data and IMU data, achieved classification accuracy ranging from 76% to 98% depending on different experimental conditions. In [7], the authors achieved a segmentation quality of 68% mIoU metric, while in [8], the mIoU on the AI4Mars dataset was 88% and 72% on the LabelMars dataset. The improvement in quality is largely achieved through the use of more complex modern neural network architectures and deep learning. It should be noted that such architectures necessarily include a proprioceptive component for assessing the immediate properties of the surface, for example, using IMU and acoustic sensors.

It is worth highlighting proprioceptive approaches that use data from IMUs and pressure sensors.

In [9], the authors proposed an open dataset consisting of nine types of indoor surfaces, with IMU data recorded as the robot moves across them. The authors compared classic machine learning methods and convolutional neural networks (CNN) and achieved a classification accuracy of approximately 64%. On the same dataset, in [10], the authors investigated the combination of convolutional layers and recurrent long short-term memory layers (LSTM) and achieved an increase in accuracy to 68%. In [11], the authors moved to using attention layers in addition to convolutional layers and LSTM layers. This allowed the authors to achieve a mean average precision of 72% on the original dataset without adding special features. This approach is further developed in [12] through optimization of the network architecture, window size, and feature engineering enabling the authors to achieve an accuracy of 97%. Meanwhile, manual feature engineering can also be quite effective when using simple machine learning methods [13]. In [14], the authors extend the approach from wheeled to legged robots, specifically testing a classifier based on IMU data processed by a recurrent neural network with gated recurrent unit (GRU). The authors achieve a very high accuracy of 99.9%. In [15], the authors examined the accuracy of a vibration-based method using an IMU for a natural outdoor environment. Despite a small set of surfaces, a total of six instances, the authors managed to achieve 98% accuracy. Furthermore, IMU can be used not only on rigid suspensions but also on spring-mounted ones; for example, in [16], the authors used a robot with shock absorbers equipped with an IMU. The acquired data were processed by a one-dimensional convolutional neural network, resulting in 80% accuracy in classifying outdoor surface types. The complexity of this approach stems from the fact that, in contrast to rigid suspensions, vibrations from the surface decay more rapidly due to damping. A specific type of vibration sensor is a conical spring with Hall effect sensors [17]. The authors proposed a novel sensor and extended the classification task by enabling the addition and identification of previously unknown surfaces. The accuracy of identifying new surfaces was 85%, and that of classifying previously known ones was 84%.

Contact or pressure sensors are most widely used for legged robots. In [18], the authors integrated data on leg position, joint velocities and accelerations, ground contact forces (measured via an embedded strain gauge), camera imagery. By processing the data using an LSTM recurrent neural network, the authors achieved a classification accuracy of approximately 96%. In [19], an acceleration sensor embedded in the leg of the robot was used to identify the contact patch via impulse duration and to classify the surface type with 97% accuracy. In [20], the authors used a test rig simulating a robot leg to obtain data from a three-axis force sensor; they processed this data using a neural network with one-dimensional convolutional layers and LSTM layers, achieving 97% accuracy.

A further classification approach relies on energy features derived from current and torque sensors, which effectively capture the energy consumption of the robot during locomotion. While this method has received comparatively little attention in surface classification, it is well established in related domains such as wheel slip detection [1, 21–23], path planning [23–27], and the characterization of terrain properties or energy consumption [28–32], suggesting considerable untapped potential for classification tasks.

In [33], the authors classified complex outdoor surfaces in Canadian forests by combining energy features with IMU data. They achieved an average accuracy of 93.9% and 93.7% with CNN and Mamba, respectively. In [34], the authors used data on motor power values and demonstrated that this information, when combined with IMU data, improves surface classification accuracy. In [35], an adaptive surface classification system based exclusively on motor currents and wheel speeds was investigated. The authors achieved accuracies of 91% for surface classification and 96% for detection of previously unknown surfaces.

The energy-based method stands as the least examined approach in this domain. Nevertheless, prior work has yielded promising accuracy when the method is combined with IMU data [33, 34] or applied independently, albeit only on a restricted indoor dataset [35]. In the present study, we evaluate the performance of the energy-based method using modern deep learning architectures and automated parameter optimization to determine whether it holds promise as a standalone classifier or functions primarily as supplementary information.

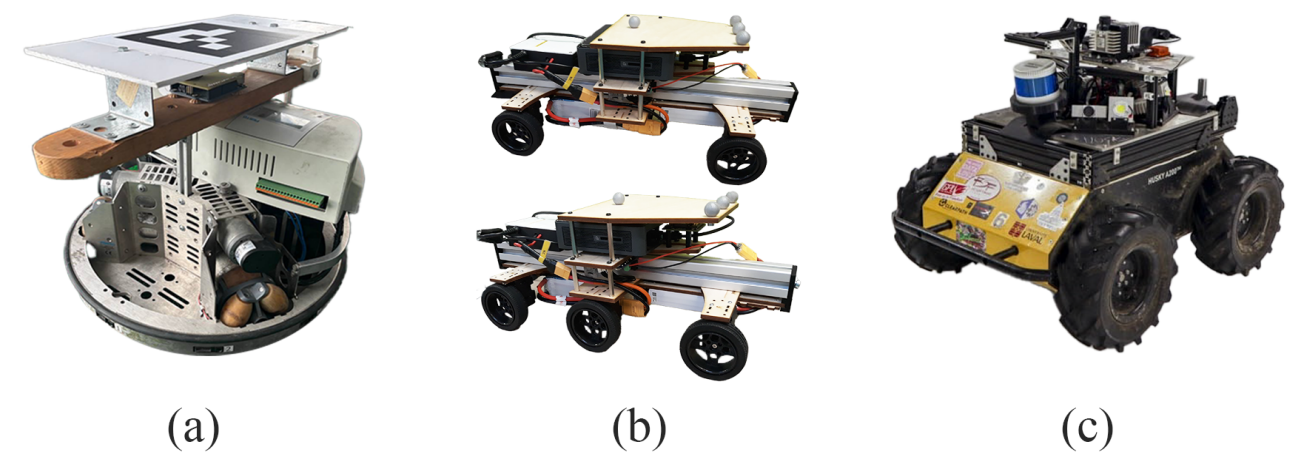

(a) (b) (c)

**Fig. 1.** Overview of robots used for data collection. (a) Omnidirectional robot for *Belyaev-Kushnarev* dataset. (b) Differential-drive robots in 4-wheel and 6-wheels configurations for *Kędzierski* dataset (c) Differential drive four-wheeled robot for *BorealTC* dataset.

## II. Method

To assess the quality of surface classification with energy features and determine whether they can function as an independent classification modality, we evaluated deep learning models trained on distinct feature sets. The composition of these sets was adapted to the sensor availability of each dataset. Specifically, we constructed energy-only feature sets derived from available energy proxies (e.g., motor currents, power estimates), IMU-only sets for datasets containing inertial data to establish a baseline, and combined sets integrating both modalities where possible.

In our work, we used four modern deep learning architectures for processing sequential data.

**Recurrent Neural Network (RNN).** We employed a recurrent architecture based on GRU cells [36]. The model structure comprises a funnel-like MLP encoder, a GRU-based recurrent block, and a linear projection layer for classification.

**Convolutional Neural Network (CNN).** The model structure consists of a series of 1D convolutional blocks and max-pooling layers [37], whose flattened output was passed to an MLP classifier. Both the number of convolutional channels and the dimensions of the classifier are organized in a funnel-like fashion, with the former expanding and the latter contracting. All convolution blocks have a kernel size of 3.

**Encoder-only Transformer.** This architecture comprises an MLP encoder, a positional encoding layer, a transformer encoder [38], and an MLP classifier. A learnable classification token is prepended to the encoded sequence to aggregate global features via the self-attention mechanism. Following the transformer blocks, the state of this token is extracted and passed to the classifier for final prediction. The MLP encoder and the classifier are organized in a funnel-like fashion, with the former expanding and the latter contracting.

**Mamba.** This architecture consists of an MLP encoder, a Mamba2 block [39], and an MLP classifier. Consistent with the other models, the encoder and classifier are organized in a funnel-like fashion, with the former expanding and the latter contracting.

## III. Experiments

Three public datasets containing data on motor currents, load torques, or energy were used for experimental studies. Two of them were collected from indoor surfaces, and one from real-world forests.

### A. Datasets

**Belyaev-Kushnarev dataset** [35]

The authors used the dataset on four types of indoor surfaces, namely: soft smooth rubber (encoded as *gray*); hard rough plastic (*green*); hard smooth laminated chipboard (*table*); medium stiffness and roughness honeycomb-shaped EVA (*red*). The authors used an omnidirectional robot, as shown in Fig. 1(a). According to the authors, the use of a holonomic platform with omni-wheels added complexity to the classification process due to the interaction of the wheels during movement. The dataset includes 504 experimental trials covering combinations of 3 target speed values (0.1, 0.2, 0.3 m/s) and 72 motion directions (from 0° to 355° in 5° steps). For each experiment, the following data were logged at 30 Hz: wheel rotational speeds ($\omega_i$) and motor currents ($I_i$) from the robot's onboard sensors, and the robot's Cartesian coordinates ($x$, $y$, $\phi$) from the computer vision system. The authors used additionally calculated metrics, primarily the energy consumption coefficient $K_e$ as follows:

$$K_e = \frac{P_{\text{real}}}{P_{\text{no-load}}} = \frac{\sum_{i=1\ldots m} \left|U_i \cdot I_i\right|}{\sum_{i=1\ldots m} \left|U_i \cdot I_{\text{no-load}}\right|} \tag{1}$$

where $P_{real}$ is the instantaneous actual power consumption at a given direction of motion, $P_{no\text{-}load}$ is the instantaneous power consumption of motion at no-load currents, m is the number of motors, $U_i$ is the motor voltage, and $I_{no\text{-}load}$ is a no-load current.

This dataset lacks IMU features, but the available energy features are present in both the raw and filtered versions. Therefore, we will also examine the effect of filtering on classification quality. We identified three sets of features to study:

1) Wheel speeds, motor currents.
2) Wheel speeds, motor currents, $K_e$.
3) Motion direction, $K_e$. This group consists of features that the authors emphasized in their study.

**Kędzierski et al. dataset** [34]

The authors used ten different surface types: long-pile carpet, artificial grass, laminate flooring, short-pile carpet, linoleum, ceramic tile, PVC foam board, OSB, foam underlayment, and EVA foam tiles. A differential-drive robot was employed in two configurations: four-wheel (4W) and six-wheel (6W). An overview of the robots is shown in Fig. 1(b). A total of 669 trials were conducted: 349 with the 4W robot and 320 with the 6W robot. The dataset contains IMU readings (linear acceleration x, y, z; angular velocity x, y, z), percentage of maximum servo torque, servo angular velocity, and the calculated average estimated power of the left and right wheel groups. The authors calculated estimated power consumption as

$$P = \tau \cdot \omega \tag{2}$$

where $\tau$ is the estimated percentage of maximum torque and $\omega$ is the angular velocity of the servomotor's shaft.

The IMU operated at 100 Hz, whereas the servos transmitted data at approximately 20 Hz. Servo data were subsequently upsampled to 100 Hz using linear interpolation to synchronize with the IMU timestamps.

We identified six sets of features to study:

1) Wheel speeds, motor currents.
2) Wheel speeds, motor currents and power.
3) Angular velocity, motor power. This set is composed by analogy with *Set 3* from the *Belyaev-Kushnarev* dataset.
4) Wheel speeds, IMU.
5) Wheel speeds, motor currents, IMU.
6) All features: Wheel speeds, motor currents and power, IMU.

**BorealTC dataset** [33]

The dataset comprises 71 trials distributed across five outdoor surface types: asphalt, flooring, ice, silty loam, and snow. A differential-drive four-wheeled robot was employed for data collection, as shown in Fig. 1(c). During data collection, the robot's speed and direction of motion were varied and not held constant across trials, meaning no standardized trajectory was enforced. Onboard sensors of the platform logged motor currents and wheel velocities at 6.5 Hz, while a synchronized IMU captured three-axis angular velocity and linear acceleration at 100 Hz. Given that the dataset includes motor currents and rotational speeds, with motor parameters obtainable from the documentation, we computed the $K_e$ parameter via (1) to assess the feature quality proposed in [35].

Similar to the *Kędzierski* datasets, we identified six sets of features to study:

1) Wheel speeds, motor currents.
2) Wheel speeds, motor currents, $K_e$.
3) Angular velocity, $K_e$.
4) Wheel speeds, IMU.
5) Wheel speeds, motor currents, IMU.
6) All features: Wheel speeds, motor currents, $K_e$, IMU

### *B. Training Workflow*

The training workflow for each dataset followed a three-stage process: first, the data was preprocessed and partitioned; second, a hyperparameter search was conducted to identify the high-performing configuration; third, the model was trained using the selected parameters.

During the data preparation stage, the original datasets were divided into large, non-overlapping segments to ensure the preservation of order within each dataset. Then, the segments were assigned to training and test subsets in a ratio of 80% to 20%, with stratification by target class. Additionally, a proxy subset was created from the training set to use in the optimization stage. The size of the segments and the specific portion of the training set used as a proxy were determined by the size and balance of the specific dataset. In the subsequent optimization and training stages, input sequences for the models were generated from each segment using a rolling window approach. To accommodate the varying feature sets and input sizes, these input sequences were generated dynamically at runtime.

The Optuna framework was used during the optimization. The proxy dataset was split into training and validation subsets, with a ratio of 80% to 20%, respectively. We conducted 80 trials per network, with each trial running for up to 20 epochs. The HyperBand algorithm [40] was used to prune unpromising trials to conserve time and resources. The optimization was performed in two stages: the optimal input sequence length was first determined for each network architecture using one representative feature set and subsequently held fixed, after which hyperparameters were tuned separately for each feature set.

In the training stage, each network was trained for an initial 150 epochs. For models that demonstrate continued improvement without reaching convergence during the first round, training was extended for an additional 150 epochs. The training process employed the AdamW optimizer [41] in conjunction with a ReduceLROnPlateau learning rate scheduler.

### *C. Optimized Parameters*

While each network architecture was designed to handle temporal sequences, they vary significantly in their internal depth and complexity. Across all models, the initial learning rate was consistently treated as an optimized hyperparameter. In addition to this training parameter, specific structural hyperparameters were optimized for each network, as detailed in the following descriptions

**Recurrent Neural Network (RNN).** The number of encoder layers, the dimension of the initial layer, and the expansion factor, along with the RNN hidden dimension, were subject to optimization. The embedding dimension was set equal to the output dimension of the final encoder layer.

**Convolutional Neural Network (CNN).** The number of layers, the initial dimension, and the expansion and contraction factors were subject to optimization.

**Encoder-only Transformer.** The number of layers, the initial dimension, and the expansion/contraction factors for both MLP modules, along with the number of transformer layers, were subject to optimization. The number of attention heads was fixed at one, while the embedding dimension was set equal to the output dimension of the final MLP encoder layer.

**Mamba.** The number of layers, the initial dimension, and the expansion/contraction factors for both MLP modules, along with the state and head dimensions of the Mamba2 block, were subject to optimization. The embedding dimension was set equal to the output dimension of the final MLP encoder layer.

TABLE I
RESULTS ON BELYAEV-KUSHNAREV DATASET

| Feature set | RNN | CNN | Mamba | Transf. | Mean |
|---|---|---|---|---|---|
| Kalman filter | | | | | |
| Set 1 | **0.9743** | **0.9661** | **0.9750** | **0.9776** | 0.9732 |
| Set 2 | **0.9797** | **0.9746** | **0.9777** | **0.9779** | 0.9775 |
| Set 3 | **0.9449** | **0.9423** | 0.9007 | **0.9334** | 0.9303 |
| No filter | | | | | |
| Set 1 | **0.9644** | **0.9692** | **0.9709** | **0.9764** | 0.9702 |
| Set 2 | **0.9729** | **0.9717** | **0.9745** | **0.9748** | 0.9735 |
| Set 3 | 0.8955 | **0.9324** | **0.9422** | 0.9008 | 0.9177 |
| Mean | 0.9553 | 0.9594 | 0.9568 | 0.9568 | |

TABLE II
RESULTS ON KĘDZIERSKI 4W DATASET

| Feature set | RNN | CNN | Mamba | Transf. | Mean |
|---|---|---|---|---|---|
| Set 1 | 0.8040 | 0.8610 | 0.7899 | 0.6447 | 0.7749 |
| Set 2 | 0.8107 | 0.8428 | 0.8101 | 0.7112 | 0.7937 |
| Set 3 | 0.8029 | 0.8499 | 0.8359 | 0.7449 | 0.8084 |
| Set 4 | **0.9733** | **0.9808** | **0.9832** | **0.9858** | 0.9808 |
| Set 5 | **0.9724** | **0.9833** | **0.9860** | **0.9801** | 0.9805 |
| Set 6 | **0.9868** | **0.9774** | **0.9843** | **0.9679** | 0.9791 |
| Mean | 0.8917 | 0.9159 | 0.8982 | 0.8391 | |

## IV. RESULTS

We investigated classification performance on three different datasets using the methods described in Chapter II with different feature sets. The results are presented in tables, where the values highlighted in bold indicate models that achieve higher accuracy compared to the best result of the authors. Classification reports for best-performing nets are presented in Appendix.

**Belyaev-Kushnarev dataset**

Input sequence length was set to 0.83 s. The authors used instantaneous values. The accuracies of the optimized and trained models are presented in table I. In the original work [35], the authors demonstrated a classification accuracy of 91%.

RNN demonstrated the best performance on *Set 2* with a Kalman filter, achieving 97.97% accuracy and improving upon the authors` results by nearly 7%. On average, CNN exhibited the highest accuracy, with other networks differing from it by no more than 0.5%. The average CNN performance surpassed that reported by the authors by nearly 5%. Across all models, the best mean performance was observed for *Set 2* with filtering enabled. Filtered data yielded accuracy comparable to unfiltered data, the average improvement was 0.65% when the filter was applied. Notably, *Set 3* underperformed the other two sets by an average of 5%. Finally, augmenting the feature set with the $K_e$ parameter alongside currents and velocities increased classification accuracy by approximately 0.5%.

**Kędzierski dataset**

We trained, analyzed, and compared the results for both robot configurations. Input sequence length was set to 1.4 s for the 4W robot and 1.5 s for the 6W robot. The authors used an input sequence length of 2 s. The accuracies of the optimized and trained models are presented in table II for the 4W configuration and in table III for the 6W configuration.

For the 4W configuration, the authors reported accuracies of 87% (IMU), 60% (estimated power), and 92% (IMU + estimated power). When using data from all sensors (*Set 6*), RNN achieved the best result of 98.68%, improving upon the authors' accuracy by 6.5%. On average, performance on *Set 6* exceeded the authors' results by nearly 6%. Using power features without IMU data, CNN trained on *Set 1* achieved the highest accuracy of 86.1%, outperforming the authors' corresponding result by 26%. On average, our models using estimated power without IMU outperformed the authors' baseline by approximately 20%. Across datasets, the highest mean accuracies were observed for *Sets 4-6*, reaching approximately 98%. Notably, models trained on *Set 3* (which includes power and direction of motion) outperformed those trained on *Sets 1-2*, a trend opposite to that observed in the *Belyaev-Kushnarev* dataset. On average, models relying solely on *Sets 1-3* (i.e., without IMU data) underperformed those incorporating IMU features by 18%, a substantial margin. However, the maximum accuracies of CNNs, for instance, were only 13% lower under these conditions, and their average accuracy remained around 85%.

For the 6W configuration, the authors reported accuracies of 88% (IMU), 74% (estimated power), and 94% (IMU + estimated power). The best-performing model on the full feature set (*Set 6*) was Mamba, with an accuracy of 99.77%, representing a 5.7% improvement over the authors' result. Meanwhile, CNN trained on *Set 5* achieved the highest overall accuracy of 99.93%. The average improvement for *Set 5* was similar, exceeding 5%. Using power data alone, CNN on *Set 1* attained the best accuracy of 99.14%, surpassing the authors' result by 25%. Overall performance trends mirrored those of the 4W configuration. Due to the uniformly high accuracies observed in the 6W configuration, where the least accurate model exceeded 97%, and the range between the worst and best models was less than 3%, it was difficult to draw definitive comparative conclusions.

Across both configurations, the use of motor power in *Set 3* yielded average accuracies 0.1-0.3% higher than those obtained using motor currents and velocities in *Sets 1-2*. Adding energy features to IMU data increased accuracy by approximately 1% on average. Conversely, models trained without IMU data performed roughly 1% worse than those incorporating inertial measurements.

**BorealTC dataset**

Input sequence length was set to 1.2 s for *Sets 1-3*, where only values from the robot's onboard system were used, and 0.8 s for *Sets 4-6*, where data from the IMU were used. The authors used an input sequence length of 1.7 s. The authors did not evaluate the accuracy of their model with only energy features, so we compared the results of our work exclusively with their accuracy on the test data of 93.96% when using CNN and IMU data. The results are shown in table IV.

TABLE III
RESULTS ON KĘDZIERSKI 6W DATASET

| Feature set | RNN | CNN | Mamba | Transf. | Mean |
|---|---|---|---|---|---|
| Set 1 | **0.9729** | **0.9914** | **0.9847** | **0.9795** | 0.9821 |
| Set 2 | **0.9730** | **0.9910** | **0.9784** | **0.9789** | 0.9803 |
| Set 3 | **0.9726** | **0.9893** | **0.9899** | **0.9804** | 0.9830 |
| Set 4 | **0.9777** | **0.9952** | **0.9949** | **0.9807** | 0.9871 |
| Set 5 | **0.9965** | **0.9993** | **0.9984** | **0.9928** | 0.9967 |
| Set 6 | **0.9933** | **0.9907** | **0.9977** | **0.9958** | 0.9944 |
| Mean | 0.9810 | 0.9928 | 0.9907 | 0.9847 | |

TABLE IV
RESULTS ON BOREALTC DATASET

| Feature set | RNN | CNN | Mamba | Transf. | Mean |
|---|---|---|---|---|---|
| Set 1 | 0.9236 | 0.9191 | 0.9140 | 0.8571 | 0.9035 |
| Set 2 | 0.9235 | 0.9183 | 0.9180 | 0.8547 | 0.9036 |
| Set 3 | 0.8408 | 0.8443 | 0.8266 | 0.7529 | 0.8161 |
| Set 4 | **0.9435** | **0.9587** | **0.9548** | 0.8997 | 0.9391 |
| Set 5 | **0.9619** | **0.9580** | **0.9621** | **0.9477** | 0.9574 |
| Set 6 | **0.9564** | **0.9678** | **0.9463** | 0.9233 | 0.9485 |
| Mean | 0.9249 | 0.9277 | 0.9203 | 0.8726 | |

CNN achieved the best result on *Set 6*, outperforming the authors' model by more than 2%. On average across models, performance on *Set 6* exceeded the authors' result by approximately 0.5%. As with the *Belyaev-Kushnarev* dataset, CNN demonstrated the highest average accuracy overall. The best result without IMU data was achieved by RNN on *Sets 1-2*. This result underperformed the authors' models that used all available data by less than 2%. The highest mean accuracy across datasets was observed on *Set 5*.

Consistent with the *Belyaev-Kushnarev* dataset, the poorest performance was observed for *Set 3* (i.e., $K_e$ and angular velocity). Models trained on *Set 3* underperformed those trained on *Set 1-2* by approximately 9%. Adding $K_e$ to the current data did not yield a significant increase in accuracy. Conversely, augmenting IMU data with current measurements improved accuracy by approximately 1.5%. On average, models lacking IMU data were 7% less accurate than those incorporating inertial measurements. However, their mean accuracy remained around 87.5%, which represents a notable result in absolute terms. Excluding *Set 3* from the comparison, the accuracy on *Set 1-2* was just over 4% lower than that of models with IMU data, with an average accuracy exceeding 90%.

Given that the behavior of *Set 3* on the *Kędzierski 4W* dataset, where it outperformed other non-IMU sets, differed from its behavior on the *Belyaev-Kushnarev* and *BorealTC* datasets, we conducted an additional analysis. Specifically, we trained supplementary models on the *Belyaev-Kushnarev* and *BorealTC* datasets using motor power and motion direction as a distinct feature set. The outcomes, however, were inconsistent. On the *Belyaev-Kushnarev* dataset, accuracy decreased by approximately 1% relative to *Set 3*, whereas on the *BorealTC* dataset, accuracy increased by 5% relative to *Set 3*, though it remained lower than the accuracies achieved on *Set 1-2*.

## V. CONCLUSION

In this work, we analyzed surface classification accuracy based on energy features across three publicly available datasets using deep learning models: RNN, CNN, encoder-only Transformer, and Mamba. Our source code and results are publicly available online: https://github.com/okushnarev/dl_surface_classification. Hyperparameters and input sequence lengths were optimized to obtain high-performing models. We achieved higher accuracy than those reported by the original authors across all three datasets. CNN exhibited the highest overall accuracy among the evaluated architectures and feature set configurations. Notably, the optimized input sequence lengths did not exceed those employed by the original authors, suggesting that our models are less data-dependent.

Using exclusively energy-based features, we attained classification accuracies in the range of 85-90%, though these results fell below those achieved with IMU-based feature sets by approximately 5-10%. We conclude that the accuracy demonstrated by classifiers relying solely on energy features supports their viability as a standalone classification approach. This is particularly compelling given the simplicity, widespread availability, and lower cost of current sensors relative to IMUs. Augmenting IMU data with energy features yielded an average accuracy improvement of approximately 1-2%, underscoring the utility of energy features for surface classification tasks.

The $K_e$ feature proposed in [35] underperformed relative to models based directly on motor currents and velocities. We attribute this to a loss of informational granularity, as neural networks appear capable of extracting more discriminative patterns from the raw energy-related signals than from the derived scalar feature.

## APPENDIX

TABLE A-I
CLASSIFICATION REPORT FOR THE BEST NET (RNN, SET 2 KALMAN) ON *BELYAEV-KUSHNAREV* DATASET

| **Surface** | **Precision** | **Recall** | **F1-score** |
|---|---|---|---|
| red | **1.0000** | **1.0000** | **1.0000** |
| gray | **0.9880** | **0.9533** | **0.9703** |
| green | **0.9539** | **0.9881** | **0.9707** |
| table | 1.0000 | **1.0000** | **1.0000** |
| macro avg | 0.9855 | 0.9854 | 0.9853 |
| weighted avg | 0.9801 | 0.9797 | 0.9797 |

TABLE A-II
CLASSIFICATION REPORT FOR THE BEST NET (CNN, SET 6) ON *BOREALTC* DATASET

| **Surface** | **Precision** | **Recall** | **F1-score** |
|---|---|---|---|
| asphalt | 0.8661 | **0.8504** | 0.8582 |
| flooring | **0.9809** | **0.9890** | **0.9849** |
| ice | **0.9926** | **0.9864** | **0.9895** |
| sandy loam | 0.9303 | **0.9749** | 0.9521 |
| snow | **0.9655** | **0.9499** | **0.9577** |
| macro avg | 0.9471 | 0.9501 | 0.9485 |
| weighted avg | 0.9679 | 0.9678 | 0.9678 |

TABLE A-III
CLASSIFICATION REPORT FOR THE BEST NET (RNN, SET 6) ON *KĘDZIERSKI 4W* (OVERALL)

| **Surface** | **Precision** | **Recall** | **F1-score** |
|---|---|---|---|
| artificial grass | **0.9909** | **0.9966** | **0.9938** |
| ceramic tiles | **0.9971** | **0.9950** | **0.9960** |
| eva foam tiles | **0.9899** | **0.9900** | **0.9900** |
| foam underlayment | **0.9813** | **0.9970** | **0.9891** |
| laminate flooring | **0.9980** | **0.9867** | **0.9923** |
| linoleum | **0.9688** | **0.9520** | **0.9603** |
| long carpet | 0.9593 | **0.9899** | **0.9744** |
| osb | **0.9930** | **0.9708** | **0.9818** |
| pvc foamboard | 0.9953 | **0.9989** | **0.9971** |
| short carpet | **0.9948** | **0.9795** | **0.9871** |
| macro avg | 0.9868 | 0.9857 | 0.9862 |
| weighted avg | 0.9869 | 0.9868 | 0.9868 |

TABLE A-IV
CLASSIFICATION REPORT FOR THE BEST NET (CNN, SET 1) ON *KĘDZIERSKI 4W* (POWER-ONLY)

| **Surface** | **Precision** | **Recall** | **F1-score** |
|---|---|---|---|
| artificial grass | **0.8098** | **0.7991** | **0.8044** |
| ceramic tiles | **0.8930** | **0.8411** | **0.8663** |
| eva foam tiles | 0.8025 | **0.8724** | **0.8360** |
| foam underlayment | **0.7919** | **0.8274** | **0.8093** |
| laminate flooring | **0.9827** | **0.9661** | **0.9743** |
| linoleum | **0.9209** | **0.8294** | **0.8727** |
| long carpet | **0.8120** | **0.8199** | **0.8159** |
| osb | **0.8669** | **0.8332** | **0.8497** |
| pvc foamboard | **0.9830** | **0.9904** | **0.9867** |
| short carpet | **0.8296** | **0.8776** | **0.8529** |
| macro avg | 0.8692 | 0.8657 | 0.8668 |
| weighted avg | 0.8627 | 0.8610 | 0.8613 |

TABLE A-V
CLASSIFICATION REPORT FOR THE BEST NET (CNN, SET 1) ON *KĘDZIERSKI 6W* (OVERALL)

| **Surface** | **Precision** | **Recall** | **F1-score** |
|---|---|---|---|
| artificial grass | **1.0000** | 1.0000 | **1.0000** |
| ceramic tiles | **0.9946** | **0.9997** | **0.9971** |
| eva foam tiles | **0.9996** | **0.9999** | **0.9997** |
| foam underlayment | **0.9999** | **1.0000** | **1.0000** |
| laminate flooring | **0.9988** | **1.0000** | **0.9994** |
| linoleum | **1.0000** | **0.9938** | **0.9969** |
| long carpet | 1.0000 | **1.0000** | **1.0000** |
| osb | **1.0000** | **0.9990** | **0.9995** |
| pvc foamboard | 1.0000 | 1.0000 | 1.0000 |
| short carpet | **1.0000** | **1.0000** | **1.0000** |
| macro avg | 0.9993 | 0.9992 | 0.9993 |
| weighted avg | 0.9993 | 0.9993 | 0.9993 |

TABLE A-VI
CLASSIFICATION REPORT FOR THE BEST NET (CNN, SET 1)
ON *KĘDZIERSKI 6W* (POWER-ONLY)

| **Surface** | **Precision** | **Recall** | **F1-score** |
| --- | --- | --- | --- |
| artificial grass | **0.9989** | **0.9807** | **0.9897** |
| ceramic tiles | **0.9843** | **1.0000** | **0.9921** |
| eva foam tiles | **0.9976** | **1.0000** | **0.9988** |
| foam underlayment | **0.9896** | **1.0000** | **0.9948** |
| laminate flooring | **0.9847** | **0.9999** | **0.9922** |
| linoleum | **1.0000** | **0.9799** | **0.9898** |
| long carpet | **0.9860** | **0.9867** | **0.9863** |
| osb | **0.9944** | **0.9899** | **0.9922** |
| pvc foamboard | **0.9955** | 1.0000 | **0.9977** |
| short carpet | **0.9808** | **0.9764** | **0.9786** |
| macro avg | 0.9912 | 0.9914 | 0.9912 |
| weighted avg | 0.9914 | 0.9914 | 0.9914 |